\documentclass[12pt,a4paper]{cibb}
\pdfoutput=1
\usepackage{fancyhdr}
\usepackage{subfigure,graphicx}
\usepackage{amsmath,amsfonts,latexsym,amssymb,euscript,xr}
\usepackage{booktabs}
\usepackage[nodayofweek]{datetime}
\usepackage{hyperref}
\usepackage[table]{xcolor}
\usepackage{color,colortbl,tabularx}
\usepackage[english]{babel}
\usepackage[protrusion=true,expansion=true]{microtype}
\usepackage{amsmath,amsfonts,amsthm}
\usepackage{pifont}
\usepackage[numbers]{natbib}

\definecolor{LightBlue}{rgb}{0.88,0.9,0.9}

\title{\Large $\ $\\ \bf Hierarchical Pretraining for Biomedical Term Embeddings}

\author{\large Bryan Cai$^{*,\dagger,1}$, Sihang Zeng$^{\dagger,2}$, Yucong Lin$^3$, Zheng Yuan$^4$, Doudou Zhou$^5$, and Lu Tian$^6$}
\address{\footnotesize $\ $\\$^1$  Department of Computer Science, Stanford University, bxcai@stanford.edu, 0000-0001-9335-5828 \\
$^2$ Department of Electronic Engineering, Tsinghua University,
Beijing, China. zengsh19@mails.tsinghua.edu.cn, 0009-0003-2921-829X \\

$^3$ Institute of Engineering Medicine, Beijing Institute of Technology, Beijing, China, linyucong@bit.edu.cn, 0000-0002-9039-0318 \\

$^4$ Alibaba Damo Academy, Hangzhou, China, yuanzheng.yuanzhen@alibaba-inc.com, 0000-0001-7179-2437 \\

$^5$  Department of Biostatistics, Harvard T.H. Chan School of Public Health, doudouzhou@hsph.harvard.edu, 0000-0002-0830-2287 \\

$^6$ Department of Biomedical Data Science, Stanford University, lutian@stanford.edu, 0000-0002-5893-0169 \\

\bigskip
$\dagger$ Equal contribution \\
$^*$corresponding author
}

\abstract{\small \textit{medical term representation, knowledge graph embedding, contrastive learning} \normalsize
\\[17pt]
{\bf Abstract.} %(max 250 words) 
Electronic health records (EHR) contain narrative notes that provide extensive details on the medical condition and management of patients. Natural language processing (NLP) of clinical notes can use observed frequencies of clinical terms as predictive features for downstream applications such as clinical decision making and patient trajectory prediction. However, due to the vast number of highly similar and related clinical concepts, a more effective modeling strategy is to represent clinical terms as semantic embeddings via representation learning and use the low dimensional embeddings as feature vectors for predictive modeling.
To achieve efficient representation, fine-tuning pretrained language models with biomedical knowledge graphs may generate better embeddings for biomedical terms than those from standard language models alone. These embeddings can effectively discriminate synonymous pairs of from those that are unrelated. However, they often fail to capture different degrees of similarity or relatedness for concepts that are hierarchical in nature.
To overcome this limitation, we propose \textsc{HiPrBERT}, a novel biomedical term representation model trained on additionally complied data that contains hierarchical structures for various biomedical terms. We modify an existing contrastive loss function to extract information from these hierarchies. Our numerical experiments demonstrate that \textsc{HiPrBERT} effectively learns the pair-wise distance from hierarchical information, resulting in a substantially more informative embeddings for further biomedical applications.
}

\begin{document}
%\thispagestyle{myheadings}
%\pagestyle{myheadings}
%\markright{\tt Proceedings of the 18th CIBB Conference 2023}%check year
% Set the page style to "fancy"...
\thispagestyle{fancy}
\pagestyle{fancy}
\fancyhead{} % clear all header fields
\renewcommand{\headrulewidth}{0pt}
\fancyhead[L]{\small \texttt{Proceedings of the 18th Conference on Computational Intelligence\\Methods for Bioinformatics \& Biostatistics (CIBB 2023)}}
\fancyfoot{} % clear all footer fields
\fancyfoot[C]{\thepage}

\section{\bf Introduction}
Biomedical term representations condense the semantic meanings of terms into a low-dimensional space, which is useful for various downstream applications, %in electronic health records (EHRs)
 such as clinical decision making %\cite{rasmy2021med}
, patient trajectory modeling %\cite{SILVA2022104195}
, and automated phenotyping. %\cite{10.1093/jamia/ocx111}. 
Current state-of-the-art methods \cite{Yuan2020CODERKI,
Zeng2022AutomaticBT,Liu2020SelfAlignmentPF} employ pretrained language models (PLMs) with contrastive learning loss to generate contextual embeddings from biomedical knowledge graphs like the Unified Medical Language System (UMLS) \cite{Bodenreider2004TheUM}. These methods focus on term normalization or entity linking problems and expect similar terms to be close in the embedding space. While they excel at similarity modeling, even in challenging tasks like unsupervised synonym grouping \cite{Zeng2022AutomaticBT}, they do not perform well in modeling hierarchies between biomedical terms \cite{kalyan_hybrid_2021}.
%This limitation is due to the nature of contrastive learning loss like multi-similarity loss \cite{msloss}, which classify samples into two categories (positive and negative) without any hierarchy. Additionally, the UMLS knowledge graph used for training does not contain any hierarchy. Thus, new training objectives and sources are necessary to encode the hierarchical information of biomedical terms in generating a better contextual embedding.

Efforts have been made in recent studies to incorporate hierarchical information into biomedical term representations. For example, Kaylan and Sangeetha (2021) used a retrofitting algorithm and UMLS relationships to incorporate ontology relationship knowledge into term representations \cite{kalyan_hybrid_2021}. However, this method treats all relationships equally. Another approach was proposed by Yang et al. (2022) based on a hierarchical triplet loss with dynamic margin learned from the hierarchy of ICD codes \cite{yang_knowledge_2022, Braemer1988InternationalSC}, which improved the performance of the ICD coding task. However, this method is less flexible as it requires an explicit parametrization of the dynamic margin, which can be difficult in the presence of many different classes of term pairs. %Additionally, the triplet loss cannot handle more than two layers of hierarchy. 
%As a result, previous methods are not suitable for creating a unified hierarchical biomedical term representation model.

To incorporate specific biomedical term hierarchies into training the embedding, we select a set of terms based on these hierarchies for each anchor term. Our model learns to improve the concordance between the cosine similarities of embedded term pairs and their similarities within hierarchies. Existing techniques for optimizing the rank loss require the specification of margins between adjacent categories\cite{%DBLP:journals/corr/SchroffKP15,
liu-etal-2022-brio}, which is delicate and time-consuming \cite{yuan2023rrhf, LeCun2006ATO}. 
%With carefully constructed hierarchies, the MS loss can be extended for rank-based optimization.

%Although ranking is a non-differentiable operator, there are existing techniques for optimizing the rank concordance \cite{Taylor2008SoftRankON, %Burges2005LearningTR,
%Cao2007LearningTR}. The triplet loss, i.e., the ranking loss, has been used in contrastive learning with more than two categories by assigning a margin between adjacent categories \cite{DBLP:journals/corr/SchroffKP15, liu-etal-2022-brio}. However, tuning the margins can be delicate and time-consuming \cite{yuan2023rrhf, LeCun2006ATO}. %and it is usually necessary to prevent the collapse of all scores being the same \cite{LeCun2006ATO}. 
%Additionally, assigned margins need to be between 0 and 1, which limits the number of categories can be considered in hierarchies. %since the cosine similarities are intrinsically bounded, which affects the model's flexibility. 
%In this paper, we tackle the ranking problem by designing a novel contrastive loss function based on the likelihood function of the conditional logistic regression, which can be modified to handle any number of ordered categories and doesn't require pre-specifying margins between categories.
%In this paper, we tackle the ranking problem by adapting an existing contrastive loss function to handle any number of ordered categories and doesn't require pre-specifying margins between categories.

In this paper, we present a novel hierarchical biomedical term representation model that leverages both the synonyms in UMLS and hierarchies in EHR codified data.  To this end, we have gathered medication terms from RxNorm \cite{Nelson2011NormalizedNF}, phenotype terms from PheCode \cite{Wu2019MappingIA}, procedure terms from CPT \cite{Dotson2013CPTCW}, and laboratory terms from LOINC \cite{McDonald2003LOINCAU}, and organize them into hierarchical structure for embedding training.  Taking advantage of constructed hierarchies, we adapt the existing contrastive loss function to handle any number of ordered categories without the need of specifying any between-category margin. We name our model \textbf{Hi}erarchical \textbf{Pr}etrained \textbf{BERT} (\textsc{HiPrBERT}).
%To address the challenge of modeling hierarchies in a contrastive learning framework, we design a novel contrastive loss function motivated from conditional logistic regression. %, which learns from the relative ranking between term pairs. 
%Our experiments demonstrate that our model effectively captures the hierarchy of biomedical terms. 
% In summary, our main contributions are:
% \begin{enumerate}
%     \item create unified representation models that are trained using both the UMLS knowledge graph and other codified EHR data that exhibit hierarchical structure.
%     \item introduce a novel contrastive loss to learn information from hierarchies.
% \end{enumerate}

\section{Related Works}
%\subsection{Biomedical Term Representations}
Biomedical term representation is the foundation of biomedical language understanding. Word embeddings generally use word2vec algorithm with biomedical corpus for training \cite{Mikolov2013DistributedRO}. Cui2vec factorizes a shifted, positive pointwise mutual information matrix to obtain a lower-dimension embedding of the words \cite{Beam2018ClinicalCE}. CODER and \textsc{SapBERT} extend the fixed vocabulary in word2vec models to arbitrary inputs by using pretrained language models and contrastive learning to learn from the synonyms in UMLS. To encode hierarchies in biomedical term representations, Yang et al. (2022) designs a hierarchical triplet loss with pre-assigned dynamic margin to learn from the hierarchy of ICD codes \cite{yang_knowledge_2022}, while Kayyan and Sangeetha (2021) uses a retrofitting algorithm to refine the representations using UMLS relationships \cite{kalyan_hybrid_2021}. These methods facilitate the development of biomedical NLP, but are still restrictive in exploring the fine information in various types of hierarchies.
%Although these methods facilitate the development of biomedical NLP, they either embed terms onto a flat space or learn from a single hierarchy. %In this paper, we introduce a unified representation model trained on the hierarchies of all common kinds of EHR codified data.
% cui2vec
% coder, sapbert
% hierarchy icd coding
% retrofit

%\subsection{Contrastive Learning Loss}
%Contrastive learning 
% nce
% infonce
% triplet loss
% multi similarity loss
% hierarch

\section{\bf Data and Methods}
\label{sec:DATA-AND-METHODS}
We will introduce the structure of the input data, the general model architecture that we use to build embeddings, the hard pair mining strategy, and the loss functions.

\subsection{UMLS and Medical Hierarchies}
\textsc{HiPrBERT} leverages two main sources of data. The first is the UMLS, a knowledge graph that encodes relations across many different medical vocabularies. These terms have no inherent order to them, and there are many different types of relations between pairs of terms. In addition to the UMLS knowledge graph, we have a collection of various hierarchies that we can leverage. Specifically, PheCODE is a hierarchy containing ICD codes that can be represented as a forest of trees. The root of each tree is a separate concept, and children of a node will represent a more specific concept. LOINC is another hierarchy representing laboratory observations, containing 171,191 nodes from 27 trees, whose depth varies from 2 to 13;  Similarly, RxNorm and CPT are also represented as forests focusing on medication and  procedure terms, respectively. PheCode contains 1,601 nodes, RxNorm contains 192,683 nodes; and CPT contains 10,360 nodes. In these hierarchies, the structure contains more information than UMLS on the ``closeness'' between various biomedical terms, which can be used to accomplish a fine embedding better discriminating closed related terms from moderately related terms. %and we use this to create more meaningful tasks. We use four different hierarchies to further train our model, with the following sizes:

%\begin{table}[ht]
%\centering
%\begin{tabular}{ll}
%Hierarchy & Nodes \\
%\hline
%PheCode & 1601 \\
%LOINC & 171191 \\
%RxNorm & 192683 \\
%CPT & 10360
%\end{tabular}
%\end{table}

It is worth noting that although the number of terms in each hierarchy is significantly lower than the number of terms in the UMLS, we expect that we can obtain enough high-quality training pairs from the hierarchy to enhance the embeddings in most relevant regions of the embedding space. In practical terms, each hierarchy consists of two mappings: one from parents to children and one from codes to the biomedical term strings.

\begin{table}[ht]
\parbox{.45\linewidth}{
\centering
\caption{Hierarchy map}
\begin{tabular}{cc}
Parent & Child \\
\hline
LP29693-6 & LP158133-1 \\
LP29693-6 & LP7798-4 \\
%LP29693-6 & LP7844-6
\end{tabular}
}
\hfill
\parbox{.45\linewidth}{
\centering
\caption{String map}
\begin{tabular}{cc}
Code & String \\
\hline
LP29693-6 & Laboratory \\
LP158133-1 & HNA \\
LP7798-4   & Fertility testing 
%LP7844-6   & Serology
\end{tabular}
}
\end{table}

\subsection{Term Embeddings}
\textsc{HiPrBERT} takes in an input term $s$ and outputs a corresponding embedding $\textbf{e}_s\in R^d$. Specifically, the input $s$ is first converted into a series of tokens, which are then encoded by HiPrBERT into a series of $d$ dimensional hidden state vectors
$$[\text{CLS}], \mathbf{t}_0, \mathbf{t}_1, ..., \mathbf{t}_n, [\text{SEP}] \xrightarrow{\textsc{HiPrBERT}}  \mathbf{h}_{[\text{CLS}]}, \mathbf{h}_0, \mathbf{h}_1, ..., \mathbf{h}_n, \mathbf{h}_{[\text{SEP}]}.$$
The embedding of $s$ is defined to be the latent vector corresponding to the [CLS] token
$$s\rightarrow \mathbf{e}_s = \mathbf{h}_{[\text{CLS}]}\in R^d.$$

%Given an input term $s$, {\color{blue} a PLM is a function mapping $s$ onto $\mathbf{e}=f(s)\in R^d$ such that the cos similarity $\mbox{cos}(\mathbf{e}_i, \mathbf{e}_j)=\mbox{cos}(f(s_i), f(s_j))$ reflects the resemblance of terms $s_i$ and $s_j$, where  $\mathbf{e}_i=f(s_i)$ and $\mathbf{e}_j=f(s_j)$ are the term embeddings for $s_i$ and $s_j$. We update the \textsc{PubMedBERT}, i.e., initial $f(\cdot)$ by minimizing the losses introduced in the next section to improve the initial embedding.}  

%{\color{red} The following description is somewhat disconnected and clear. I have tried to rewrite it. Please see above and edit as you see fit. LT }

\subsection{Distance metric}
Similar to \textsc{SapBERT}, our approach learns term representations by maximizing the embedding similarity between term-term pairs that are ``close" and minimizing embedding similarities between term-term pairs that are ``far". We define the embedding similarity between terms $s_i$ and $s_j$ as $S_{ij} = \cos(\mathbf{e}_i, \mathbf{e}_j).$ We also define following distances to quantify the resemblance between terms $s_i$ and $s_j.$ These particular choices of the numerical value are not important and only their order matters in training embeddings.  

\begin{enumerate}
\item If $s_i$ and $s_j$ are from the UMLS, 
$d(s_i, s_j) = 
\begin{cases}
0  & \text{$s_i$ and $s_j$ are synonyms;} \\
3  & \text{otherwise.}
\end{cases}$
\item If $s_i$ and $s_j$ are from a hierarchy, 
$d(s_i, s_j) = 
\begin{cases}
0  & \text{$s_i$ and $s_j$ are synonyms;} \\
1  & \text{$s_i$ and $s_j$ have the same parent (a sibling pair);} \\
2  & \text{$s_i$ and $s_j$ are a parent-child pair;} \\
3  & \text{otherwise.}
\end{cases}$
\end{enumerate}

\subsection{Hard Pair Mining}
When sampling UMLS term data, we use an online triplet miner to select negative pairs. Specifically, among all triplets of terms $(s_a, s_p, s_n)$, where $(s_a, s_p)$ are synonymous and $(s_a, s_n)$ are non-synonymous, based on initial embeddings, $(s_a, s_p, s_n)\rightarrow (\mathbf{e}_a, \mathbf{e}_p, \mathbf{e}_n)$, we consider the difference between $ \mbox{cos}(\mathbf{e}_a, \mathbf{e}_p)$ and $\mbox{cos}(\mathbf{e}_a, \mathbf{e}_n),$ and select the triplets with this difference $>0.25$ to be included in our minibatch for further training. We do the same for UMLS relational data.

%the triplet loss is defined as
%\begin{equation*}
%\mathcal{L}_{triplet} := [S_{an} - S_{ap} + \lambda_m]_+
%\end{equation*}
%where $[x]_+=xI(x>0)$ and $\lambda_m = 0.2$. We then select triplets that violate the margin, i.e. have positive triplet loss, to include in our minibatch. We can do the same for UMLS relational data.

For hierarchical data, we leverage the structure of the tree to construct minibatches. For example, we use distance 0 pairs as positive samples, and distance $>0$ pairs as negative samples. We do this with every distance to encourage separation between varying levels of similarity.

\subsection{Loss Function}
Given an anchor term $s_i$ and a set of terms $\Omega_i$, we can define the sets 
$$
\Omega_i^{(0)}(d_0) = \left\{j \in \Omega_i \mid d(s_i,s_j) \le d_0 \right\} ~~\mbox{and}~~\Omega_i^{(1)}(d_0) = \left\{j \in \Omega_i \mid d(s_i,s_j) > d_0 \right\}. 
$$
In other words, $\Omega_i^{(0)}(d_0)$ contains all terms that are at most distance $d_0$ from $s_i$, $\Omega_i^{(1)}(d_0)$ contains all terms that are further than $d_0$ away. Our goal is to create embeddings such that the similarity between $s_i$ and terms in $\Omega_i^{(0)}(d_0)$ is greater than that between $s_i$ and terms in $\Omega_i^{(1)}(d_0)$. We use the multi-similarity loss \cite{msloss}. %, using the hard pair mining strategy from above. 
For UMLS data, we have the standard MS loss function.
\begin{equation*}
\sum_{i=1}^{k}\left[\alpha^{-1}\log \left(1+\sum_{j \in \Omega_i^{(0)}(0)} e^{-\alpha\left(S_{ij}-\lambda\right)}\right) +\beta^{-1}\log \left(1+\sum_{j \in \Omega_i^{(1)}(0)} e^{\beta\left(S_{ij}-\lambda\right)}\right)\right],
\end{equation*}
where $\alpha=2, \beta=2, \lambda=.5$. Note that the terms in $\Omega_i^{(1)}(0)$ come from the triplet mining procedure. For hierarchical data, we use a modified loss:
\begin{equation*}
%\mathcal{L}_{M S}=&
\sum_{d_0=0}^{2} \sum_{i=1}^{k}\left[\alpha^{-1}\log \left(1+\sum_{j \in \Omega_i^{(0)}(d_0)} e^{-\alpha\left(S_{ij}-\lambda\right)}\right) +\beta^{-1}\log \left(1+\sum_{j \in \Omega_i^{(1)}(d_0)} e^{\beta\left(S_{ij}-\lambda\right)}\right)\right],
\end{equation*}
with the same set of tuning parameters.

\section{Experiments}
\subsection{Model Training}
Our training process is similar to that of \textsc{SapBERT}, with the main key difference being the loss functions that were used. Using PyTorch \cite{Paszke_PyTorch_An_Imperative_2019} and the transformers library \cite{Wolf_Transformers_State-of-the-Art_Natural_2020}, our model was initialized from \textsc{PubMedBERT} \cite{gu2020pubmedbert} and trained using AdamW \cite{loshchilov2019} with a learning rate of $2\times 10^{-5}$, a weight decay rate of 0.01, and linear learning rate scheduler. We use a training batch size of 256, and train on the preprocessed UMLS synonym data, UMLS relation data, and hierachical data for one epoch. This equates to about 120 thousand iterations, and takes less than 10 hours on a single GPU machine.

\subsection{Model Evaluation}
To objectively evaluate our models, we randomly selected evaluation pairs from hierarchies that were not used in model training. For each evaluation pair, we calculated the cosine similarity between the respective embeddings to determine their relatedness. The quality of the embedding was measured using the AUC under the ROC curve for discriminating between distance $i$ pairs and distance $j$ pairs, where $0\le i<j\le 3$. In addition, we have also evaluated the embedding performance via Spearman's correlation and precision-recall curve.

For relatedness tasks, we used pairs of terms in our holdout set for various relations to test the models. There are many different types of relationships, and we report three of clinical importance, as well as the average of the 28 most common relations. We also included performance on the Cadec term normalization task.

We compare \textsc{HiPrBERT} with a set of competitors including \textsc{SapBERT}, CODER, \textsc{PubMedBERT}, \textsc{BioBERT}, \textsc{BioGPT} and \textsc{DistilBERT}, where the \textsc{SapBERT} is retrained without using testing data for generating fair comparisons. 

\subsection{Evaluation Results}
%\subsection{Tree Distance Discrimination}
%Our first task aims to gauge how well a model can distinguish between pairs of terms that are more strongly related and pairs of terms that are less related. To achieve this, we can select pairs of term with distance 1 and 2 from our evaluation set, and use the cosine similarity between the respective embeddings to classify the relatedness of the pairs. We can to compute the AUC of this classifier to measure how well the embeddings discriminate between the two types of pairs. We can repeat this with other pairs of distances as well.
The AUC values for discriminating pairs of different distances are reported in Table \ref{table:auc1}. \textsc{HiPrBERT}, fine-tuned on hierarchical datasets, outperforms all its competitors in every category, except for 1 vs 3, where it's performance is very close to CODER. The most noteworthy improvement is in the 0 vs 1 task, where models have to distinguish synonyms from very closely related pairs, such as ``Type 1 Diabetes'' and ``Type 2 Diabetes''. We have also reported the results using Spearman's rank correlation coefficient in Table \ref{table:spear1}, and the conclusions are similar.
%The model using MCL loss performs slightly better in discriminating 1 vs 2 and 1 vs 3, as compared to the simple CL loss which only considers contrast between 0 and $>0$.  For cadec, while our models do not match the performance of the published CODER, which has been optimized and fine-tuned, they still perform favorably. The rank correlations between cosine similarity and expert-curated similarity show similar pattern. As expected, the MCL-based embedding outperforms the other three trained models by leveraging additional information from hierarchies. MCL-based embedding also demonstrates a substantially improved performance in comparison with the initial \textSC{BioBERT} representation on all metrics.

We also see significant improvements in all relatedness tasks (Table \ref{table:auc2}). For example, the AUC in the ``Causative'' category improves from 91.9\% to 98.1\% in comparison with the second best embedding generated by CODER. Similar improvement has been also observed in detecting ``May Cause/Treat" and ``Method of" relations. Overall, the average performance of the model in detecting the 28 most common relationships improved from 88.6\% to 93.7\% in comparison with the next best embedding. This demonstrates a substantial improvement in our ability to capture more nuanced information. It is worth noting that \textsc{HiPrBERT}'s performance in Cadec is on par with other existing models, indicating that our model does not compromise on performance in similarity tasks while achieving improvements in other areas. Lastly, the comparison results based on Spearman's correlation (Table \ref{table:spear1}) and precision-recall curve (not reported) are similar. 

\section{Discussion}

Our model is one of the first to include terms from medical term hierarchies (PheCODE, LOINC, RxNorm), and these trees contain terms critical for structured EHR data. Existing methods such as \textsc{CODER} and \textsc{SapBERT} do not train on this specific vocabulary. By improving embeddings for these strings in particular, our embeddings have the potential to integrate better with structured EHR data, enhancing the representation of patients. 
This then directly leads improvements in downstream tasks such as extracting prediction features and patients clustering.

The use of induced distance from hierarchies helps improve model performance, and can be expanded in several ways. One may consider more pair types within each hierarchy; for example the distance metric can be expanded to include grandparent-child and uncle-nephew pairs. Alternatively, the distance metric can take into account the global structure of the tree. Currently, pairwise resemblance only takes into account the local information around the term, looking only at immediate connections. However, typically nodes closer to the root of the hierarchies represent broader concepts that are further apart, whereas nodes closer to the leaves represent more specific concepts that are closer together. This can either be explicitly coded into the training process, or ideally learnt on the fly. In addition, different hierarchies will naturally differ in structure and therefore pairwise distance, so this adjustment would be hierarchy specific. Our simple choice here is for computational convenience and can be improved.

\section{Conclusion}
In this paper we present a novel method for training embeddings better discriminating pairs of different similarity by taking advantage of additional hierarchical structures. Operationally, the method only requires to order the term-term similarity, which is much simpler than assigning quantitative margins between similarities used in the rank loss.   
%present in addit trees, we can construct informative positive and negative pairs to achieve a more fine grained ordering of cosine similarity. 
The new model outperforms existing ones on separating weakly related terms from closely related terms without sacrificing performance on other metrics. 

\begin{table}[h]
\centering
\caption{Tree Results (ROC AUC)}
\label{table:auc1}
\begin{tabular}{lllllll}
Model & \multicolumn{6}{c}{Distance Categories}\\
\cline{2-7}
& (0, 1) & (0, 2) & (0, 3) & (1, 2) & (1, 3) & (2, 3)\\
\hline
\textsc{SapBERT}$^0$ & 0.636 & 0.779 & 0.967 & 0.702 & 0.971 & 0.905 \\
\textsc{CODER} & 0.599 & 0.737 & 0.977 & 0.679 & \textbf{0.979} & 0.931 \\
\textsc{PubMedBERT}$^1$ & 0.539 & 0.609 & 0.744 & 0.575 & 0.721 & 0.653 \\
\textsc{BioBERT} & 0.497 & 0.576 & 0.599 & 0.586 & 0.611 & 0.523 \\
\textsc{BioGPT} & 0.571 & 0.667 & 0.795 & 0.604 & 0.754 & 0.662 \\
\textsc{DistilBERT} & 0.544 & 0.631 & 0.729 & 0.589 & 0.694 & 0.614  \\
\hline
\textsc{HiPrBERT} & \textbf{0.657} & \textbf{0.796} & \textbf{0.986} & \textbf{0.704} & 0.977 & \textbf{0.936} \\
\end{tabular}
\\
\footnotesize{$^0:$ Representation trained after removing evaluation data $^1:$ Initial representation for our model training}
\end{table}

\begin{table}[ht]
\centering
\caption{Tree Results (Spearman's Correlation)}
\label{table:spear1}
\begin{tabular}{lllllll}
Model & \multicolumn{6}{c}{Distance Categories}\\
\cline{2-7}
& (0, 1) & (0, 2) & (0, 3) & (1, 2) & (1, 3) & (2, 3)\\
\hline
\textsc{SapBERT} & 0.198 & 0.483 & 0.800 & 0.294 & 0.627 & 0.693 \\
\textsc{CODER} & 0.144 & 0.411 & 0.817 & 0.262 & \textbf{0.638} & 0.738 \\
\textsc{PubMedBERT} & 0.057 & 0.190 & 0.417 & 0.109 & 0.294 & 0.262 \\
\textsc{BioBERT} & -0.004 & 0.131 & 0.169 & 0.125 & 0.148 & 0.039 \\
\textsc{BioGPT} & 0.104 & 0.289 & 0.506 & 0.152 & 0.338 & 0.277 \\
\textsc{DistilBERT} & 0.065 & 0.227 & 0.393 & 0.130 & 0.259 & 0.196  \\
\hline
\textsc{HiPrBERT} & \textbf{0.229} & \textbf{0.512} & \textbf{0.832} & \textbf{0.298} & 0.635 & \textbf{0.747} \\
\end{tabular}
\end{table}

\begin{table}[h]
\centering
\caption{Other Tasks}
\label{table:auc2}
\begin{tabular}{llllll}
Model & \multicolumn{4}{c}{Relatedness Tasks} & \\
\cline{2-5}
& Causative & May Cause/Treat & Method of & Top Rel.$^0$ & Cadec (Top 1/3) \\
\hline
\textsc{SapBERT} & 0.889 & 0.791 & 0.880 & 0.869 & \textbf{0.610}/0.801 \\
\textsc{CODER} & 0.919 & 0.750 & 0.902 & 0.886 & 0.585/0.760 \\
\textsc{PubMedBERT} & 0.711 & 0.706 & 0.554 & 0.572 & 0.172/0.238 \\
\textsc{BioBERT} & 0.508 & 0.499 & 0.503 & 0.534 & 0.083/0.116 \\
\textsc{BioGPT} & 0.771 & 0.687 & 0.582 & 0.612 & 0.221/0.299 \\
\textsc{DistilBERT} & 0.683 & 0.605 & 0.472 & 0.579 & 0.177/0.245 \\
\hline
\textsc{HiPrBERT} & \textbf{0.981} & \textbf{0.803} & \textbf{0.923} & \textbf{0.937} & 0.605/\textbf{0.805} \\
\end{tabular}
\\
\footnotesize{$^0:$ Average performance over 28 most common relations in UMLS}
\end{table}

\footnotesize
\bibliographystyle{IEEEtranN}
\bibliography{main} 
\normalsize

\end{document}